\documentclass{article}

\usepackage{archive/neurips_2025}

\usepackage[utf8]{inputenc} 
\usepackage[T1]{fontenc}    
\usepackage{hyperref}       
\usepackage{url}            
\usepackage{booktabs}       
\usepackage{amsfonts}       
\usepackage{amsmath,amssymb}
\usepackage{microtype}      
\usepackage{graphicx}
\usepackage{enumitem}
\usepackage{wrapfig}
\usepackage{xcolor}
\usepackage{graphicx}
\usepackage{algorithm}
\usepackage{algorithmic}
\begin{document}

\title{Counterfactual Self-Questioning for Stable Policy Optimization in Language Models}]

\author{%
  Mandar Parab \\
  \texttt{mandar4tech@gmail.com}
}

\maketitle

\begin{abstract}
Recent advances in language model self-improvement, including self-reflection \cite{reflexion2023}, 
step-wise verification \cite{cove2023,wei2023letsverify}, debate \cite{du2023debate}, and 
self-reward optimization \cite{feng2025selfrewarding}, demonstrate that models can iteratively 
refine their own reasoning. However, these approaches typically depend on external critics, 
hand-crafted reward models, or ensemble sampling, introducing additional supervision and 
instability during training.

We propose \textbf{Counterfactual Self-Questioning (CSQ)}, a framework in which a single language 
model generates counterfactual critiques of its own reasoning and uses these internally generated 
trajectories as a structured policy optimization signal. CSQ decomposes learning into three stages: 
(1) an initial policy rollout producing a base reasoning trajectory; (2) self-questioning, where 
the model formulates targeted counterfactual probes conditioned on its own reasoning; and 
(3) counterfactual critique, where alternative trajectories expose faulty assumptions, missing 
constraints, or invalid steps.

The resulting counterfactual trajectories provide relative feedback that can be directly integrated 
with policy optimization methods such as Group Relative Policy Optimization (GRPO) \cite{grpo2023}, 
without requiring external reward models or multiple agents. Across GSM8K \cite{cobbe2021gsm8k}, 
MATH \cite{hendrycks2021math}, and Minerva-style quantitative reasoning tasks 
\cite{lewkowycz2022minerva}, CSQ improves accuracy by \textbf{+6.7 to +12.4 points} over standard 
chain-of-thought prompting \cite{wei2022cot} and by \textbf{+3.1 to +5.8 points} over strong 
verification-based baselines. Ablation studies show that counterfactual self-questioning yields 
more diverse failure discovery, more precise error localization, and more stable training dynamics 
than prior self-improvement methods such as STaR \cite{zelikman2022star}, Self-Discover 
\cite{madaan2024selfdiscover}, and Self-Rewarding Language Models \cite{feng2025selfrewarding}.

These results suggest that counterfactual self-questioning provides a scalable and stable alternative 
to external critics for policy optimization in language models, enabling robust reasoning improvement 
using internally generated training signals.
\end{abstract}

\section{Introduction}

Large language models (LLMs) have achieved strong performance on mathematical and logical
reasoning tasks when equipped with structured prompting techniques such as
chain-of-thought \cite{wei2022cot}, step-wise verification \cite{wei2023letsverify}, and
domain-specific training \cite{lewkowycz2022minerva}. Despite these advances, LLM
reasoning remains brittle: small errors in intermediate steps often propagate, models
exhibit overconfident hallucinations, and failures are difficult to detect without
external verification \cite{cove2023}. Improving reasoning reliability therefore
requires mechanisms that expose and correct internal failure modes rather than relying
solely on final-answer supervision.

Recent work explores whether LLMs can improve themselves through internally generated
feedback. Approaches such as Reflexion \cite{reflexion2023}, STaR \cite{zelikman2022star},
Self-Discover \cite{madaan2024selfdiscover}, debate \cite{du2023debate}, and
self-rewarding language models \cite{feng2025selfrewarding} demonstrate that models can
iteratively refine their reasoning. However, these methods typically rely on external
critics, multi-agent debate, extensive sampling, or auxiliary verifier models, increasing
computational cost and architectural complexity.

In contrast, human reasoning often relies on targeted counterfactual interrogation such as 
asking whether a particular step might be wrong and exploring the consequences before
committing to a conclusion. This suggests an alternative paradigm for LLM
self-improvement one based on internally generated counterfactual critique rather than
external verification.

In this paper, we introduce \textbf{Counterfactual Self-Questioning}, a framework in
which a single language model generates and evaluates counterfactual critiques of its
own reasoning. Given an initial chain-of-thought solution, the model produces targeted
``What if this step is wrong?'' probes, simulates alternative reasoning trajectories, and
uses the resulting signals to refine its policy. Counterfactual critiques are generated
by lightweight ego critics that share parameters with the base model and introduce no
additional learned components.

Our approach differs from prior self-improvement methods in three key ways.
First, critique is generated from a single policy rollout rather than from ensembles,
external critics, or stored successful trajectories. Second, counterfactual reasoning is
applied within the model’s own reasoning trajectory rather than at the input or data
level. Third, the resulting critiques are converted into structured learning signals
using Group Relative Policy Optimization (GRPO) \cite{grpo2023}, enabling stable policy
updates without a learned value function.

We evaluate Counterfactual Self-Questioning on established mathematical reasoning
benchmarks, including GSM8K \cite{cobbe2021gsm8k}, MATH \cite{hendrycks2021math}, and
Minerva-style quantitative reasoning tasks \cite{lewkowycz2022minerva}. Across four model
families and multiple capacity regimes, the proposed method improves accuracy over
standard chain-of-thought baselines, with the largest gains observed for small and
medium-sized models. Ablation studies show that one or two counterfactual critics provide
the best balance between critique diversity and optimization stability.

In summary, this work makes the following contributions:
\begin{itemize}[leftmargin=1.1em]
    \item We propose Counterfactual Self-Questioning, a verifier-free framework for
    improving LLM reasoning via internally generated counterfactual critique.
    \item We introduce a simple training and inference pipeline that converts
    counterfactual critiques into structured policy optimization signals using GRPO.
    \item We demonstrate consistent improvements on GSM8K, MATH, and Minerva-style tasks
    across multiple model sizes, with detailed analysis of stability and scaling behavior.
\end{itemize}

\section{Related Work}

Our work relates to prior efforts on improving language model reasoning through
self-improvement, verification, multi-agent feedback, counterfactual reasoning, and
reinforcement learning with model-generated signals. We position
\emph{Counterfactual Self-Questioning} as a method for constructing an internal,
trajectory-level policy optimization signal that complements existing approaches.

\paragraph{Self-Improvement and Iterative Reasoning:}
Several methods explore whether language models can improve their own reasoning using
internally generated feedback. Reflexion \cite{reflexion2023} introduces memory-based
self-correction, STaR \cite{zelikman2022star} bootstraps improved policies from
model-generated correct solutions, and Self-Discover \cite{madaan2024selfdiscover}
synthesizes new reasoning strategies through internal feedback. Self-consistency
sampling \cite{wang2022selfconsistency} reduces variance by aggregating multiple
reasoning paths. These approaches typically rely on extensive sampling, replay buffers,
or external filtering. In contrast, Counterfactual Self-Questioning generates critique
from a single policy rollout by probing alternative counterfactual trajectories,
avoiding reliance on large ensembles or stored solutions.

\paragraph{Verification, Critics, and Debate:}
Another line of work reduces hallucinations through explicit verification.
Chain-of-Verification (CoVe) \cite{cove2023} and step-wise verification
\cite{wei2023letsverify} validate intermediate reasoning, often using separate verifier
models. Debate-based methods \cite{du2023debate} and model-based critics
\cite{akopyan2024critic,zheng2023llmjudge} expose errors through adversarial interaction,
while Constitutional AI \cite{bai2022constitutional} uses predefined principles to guide
self-critique. These approaches introduce additional models, agents, or rules. By
contrast, Counterfactual Self-Questioning distills critique into a single-policy setting
where counterfactual probes act as an implicit internal opponent without auxiliary
components.

\paragraph{Counterfactual Reasoning:}
Counterfactual reasoning has been widely used to improve robustness and causal
generalization in NLP. Counterfactual data augmentation encourages models to capture
causal structure rather than spurious correlations \cite{xu2021counterfactual}.
Counterfactual thinking also underlies logical reasoning tasks that require evaluating
alternative hypotheses. Our work differs in applying counterfactual reasoning
within the model’s own reasoning trajectory rather than at the input or data
level, enabling introspective error discovery during inference and training.

\paragraph{Reinforcement Learning and Self-Generated Rewards:}
Reinforcement learning from human feedback (RLHF) \cite{ouyang2022instructgpt} and policy
optimization methods such as PPO \cite{schulman2017ppo} are central to modern LLM
training. Recent work shows that models can generate their own reward signals
\cite{feng2025selfrewarding}, while Group Relative Policy Optimization (GRPO)
\cite{grpo2023} provides a stable framework for learning from relative feedback.
Counterfactual Self-Questioning complements this line of work by producing structured,
trajectory-level feedback that can be directly integrated into GRPO-style optimization.

\paragraph{Evaluation Benchmarks:}
Mathematical reasoning benchmarks such as GSM8K \cite{cobbe2021gsm8k}, MATH
\cite{hendrycks2021math}, and Minerva-style datasets \cite{lewkowycz2022minerva} are
standard for evaluating reasoning quality. While scaling laws \cite{kaplan2020scaling}
highlight the role of model capacity, structured reasoning methods such as
chain-of-thought prompting \cite{wei2022cot} demonstrate that reasoning strategy and
error mitigation are equally important. Our evaluation follows this established protocol.

Across prior work, the use of internally generated counterfactual probes as a unified
learning signal remains underexplored. Existing methods emphasize reflection,
verification, debate, or reward modeling, but do not systematically generate and resolve
counterfactual alternatives to the model’s own reasoning trajectory. Counterfactual
Self-Questioning fills this gap by introducing a single-policy mechanism that produces
structured counterfactual feedback usable for both inference and policy optimization,
offering a lightweight and scalable alternative to external critics.

\section{Methodology}

\begin{figure}[t]
    \centering
    \includegraphics[width=\linewidth]{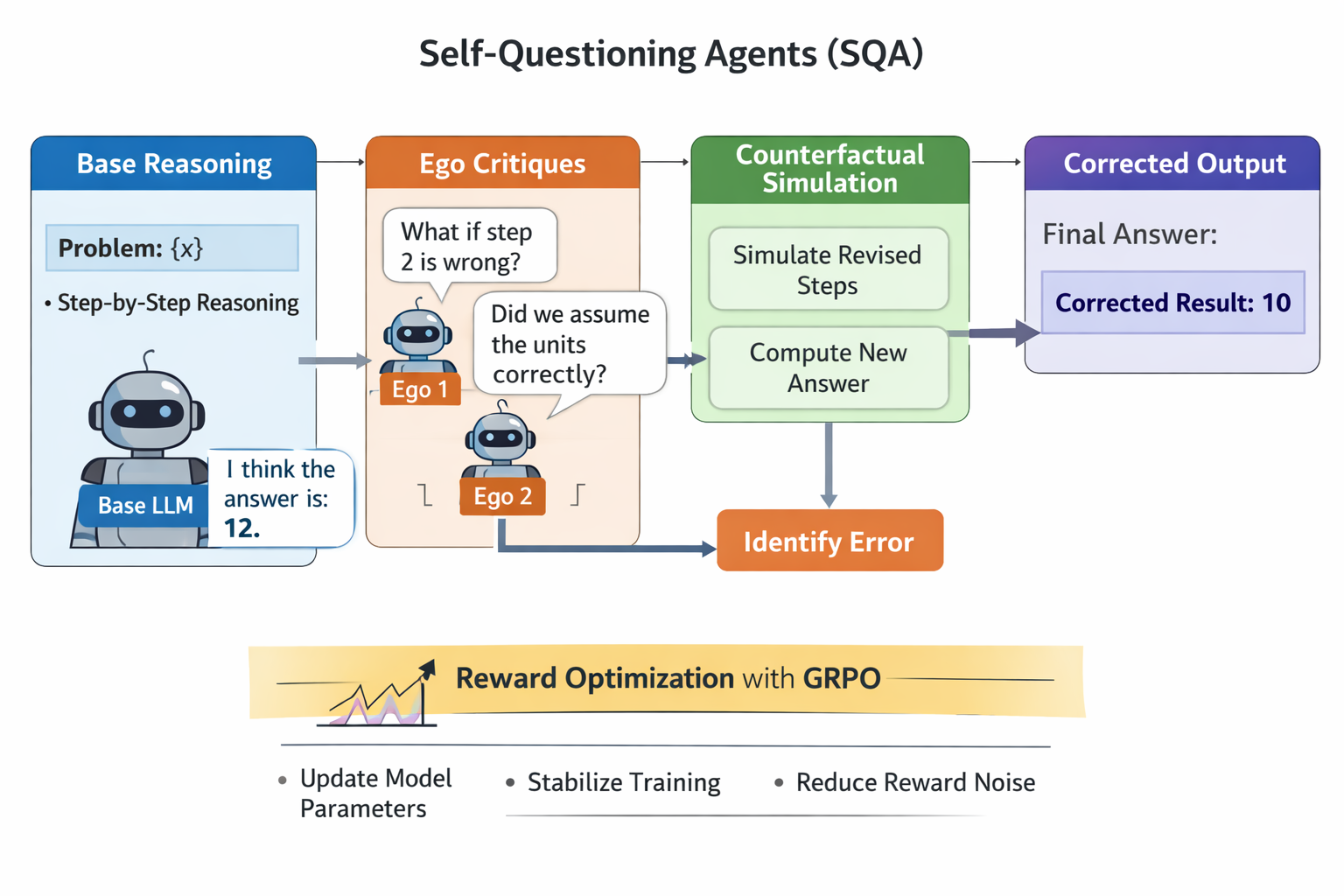}
    \caption{Overview of Counterfactual Self-Questioning (CSQ). 
    A base policy generates an initial reasoning trajectory. Counterfactual
    self-questioning produces alternative trajectories that expose failure modes.
    These trajectories are used as relative feedback for policy optimization via GRPO.}
    \label{fig:csq_overview}
\end{figure}

We propose \textbf{Counterfactual Self-Questioning (CSQ)}, a training and inference
framework in which a single language model generates and evaluates counterfactual
critiques of its own reasoning. CSQ does not rely on external critics, auxiliary reward
models, or multi-agent debate. Instead, it constructs a structured policy optimization
signal by comparing a base reasoning trajectory with internally generated
counterfactual alternatives. Figure~\ref{fig:csq_overview} illustrates the overall
workflow.

\subsection{Problem Setup}

We consider a dataset of reasoning problems
\[
\mathcal{D} = \{(x_i, y_i)\},
\]
where $x_i$ denotes an input problem and $y_i$ its ground-truth answer. Let
$\pi_\theta$ denote a language model policy parameterized by $\theta$. Given an input
$x$, the policy generates a reasoning trajectory followed by a final answer:
\[
\tau^{(0)} \sim \pi_\theta(\cdot \mid x), \qquad \hat{y}^{(0)} = f(\tau^{(0)}).
\]
This base trajectory corresponds to standard chain-of-thought reasoning and is
unverified.

Our goal is to improve $\pi_\theta$ by enabling it to identify and correct failures in
its own reasoning trajectory through internally generated counterfactual feedback.
Crucially, all components of CSQ share the same parameters $\theta$; counterfactual
generation introduces no additional models or learnable parameters.

\subsection{Counterfactual Self-Questioning}

Given the base trajectory $\tau^{(0)}$, CSQ generates a set of counterfactual probes that
target potential failure points in the reasoning process. Each probe is conditioned on
the base trajectory and prompts the model to consider an alternative hypothesis, such
as the possibility of an incorrect intermediate step or a missing constraint.

Formally, for each input $x$, we generate $N_{\text{cf}}$ counterfactual trajectories:
\[
\tau^{(k)} \sim \pi_\theta(\cdot \mid x, \tau^{(0)}, q^{(k)}), \qquad k = 1,\dots,N_{\text{cf}},
\]
where $q^{(k)}$ denotes a counterfactual query derived from the base reasoning. These
queries encourage the model to revise assumptions, re-evaluate computations, or explore
alternative solution paths. Counterfactual trajectories are explicitly instructed to
expose and repair potential errors rather than produce arbitrary disagreement.

Each trajectory $\tau^{(k)}$ yields a candidate answer $\hat{y}^{(k)}$. At inference
time, the set $\{\tau^{(0)}, \tau^{(1)}, \dots, \tau^{(N_{\text{cf}})}\}$ can be used as a
lightweight verification mechanism. During training, these trajectories form a
comparison group for policy optimization.

\subsection{Policy Optimization with GRPO}
\label{sec:grpo-training}

We optimize $\pi_\theta$ using Group Relative Policy Optimization (GRPO)
\cite{grpo2023}, which is well suited for settings where multiple trajectories are
available for the same input. For each problem $x$, we define a trajectory group
\[
\mathcal{G}(x) = \{\tau^{(0)}, \tau^{(1)}, \dots, \tau^{(N_{\text{cf}})}\}.
\]

Each trajectory $\tau \in \mathcal{G}(x)$ receives a scalar reward
\[
R(\tau) = \alpha R_{\text{correct}}(\tau) + \beta R_{\text{repair}}(\tau)
           - \gamma R_{\text{instability}}(\tau),
\]
where $R_{\text{correct}}$ indicates answer correctness,
$R_{\text{repair}}$ rewards trajectories that correct errors present in the base
trajectory, and $R_{\text{instability}}$ penalizes incoherent or internally inconsistent
counterfactual reasoning.

GRPO computes a group-level baseline
\[
b(x) = \frac{1}{|\mathcal{G}(x)|} \sum_{\tau \in \mathcal{G}(x)} R(\tau),
\]
and defines the relative advantage
\[
A(\tau) = R(\tau) - b(x).
\]
The policy update is given by
\[
\theta \leftarrow \theta + \eta
\mathbb{E}_{\tau \sim \mathcal{G}(x)}
\left[\nabla_\theta \log \pi_\theta(\tau \mid x)\, A(\tau)\right].
\]

Because all trajectories are generated by the same policy, optimization internalizes
the corrective patterns exposed by counterfactual reasoning. Over training, the base
policy increasingly produces more reliable reasoning trajectories without requiring
explicit counterfactual prompting at inference time.

\begin{algorithm}[t]
\caption{Counterfactual Self-Questioning (CSQ)}
\label{alg:csq}
\begin{algorithmic}[1]
\REQUIRE Dataset $\mathcal{D} = \{(x_i, y_i)\}$, policy $\pi_\theta$, 
number of counterfactuals $N_{\text{cf}}$
\FOR{each training example $(x, y) \in \mathcal{D}$}
    \STATE Generate base reasoning trajectory $\tau^{(0)} \sim \pi_\theta(\cdot \mid x)$
    \STATE Extract base answer $\hat{y}^{(0)}$
    \FOR{$k = 1$ to $N_{\text{cf}}$}
        \STATE Generate counterfactual query $q^{(k)}$ conditioned on $\tau^{(0)}$
        \STATE Generate counterfactual trajectory 
        $\tau^{(k)} \sim \pi_\theta(\cdot \mid x, \tau^{(0)}, q^{(k)})$
    \ENDFOR
    \STATE Form trajectory group $\mathcal{G}(x) = \{\tau^{(0)}, \dots, \tau^{(N_{\text{cf}})}\}$
    \STATE Compute rewards $R(\tau)$ for all $\tau \in \mathcal{G}(x)$
    \STATE Update $\theta$ using GRPO over $\mathcal{G}(x)$
\ENDFOR
\end{algorithmic}
\end{algorithm}

\subsection{Implementation Details}

In practice, we use a small number of counterfactual trajectories,
$N_{\text{cf}} \in \{1,2,3\}$, which balances critique diversity and optimization
stability. A single counterfactual trajectory often identifies isolated arithmetic
errors, while two trajectories reliably expose complementary failure modes such as
incorrect assumptions and missing constraints. Larger values introduce redundancy and
noise, yielding diminishing returns.

We evaluate CSQ across models of varying capacity, including
\texttt{Llama-3.2-1B-Instruct}, \texttt{Llama-3.2-3B-Instruct},
\texttt{Qwen2-0.5B-Instruct}, and \texttt{Mathstral-7B-v0.1}. Training is performed with a
learning rate of $1\times10^{-6}$ for 3--5 epochs, batch size 4 with gradient
accumulation of 2, and counterfactual generations capped at 256 tokens.

Across models, CSQ consistently improves accuracy by 2--10 absolute points, with the
largest relative gains observed for smaller models. These results indicate that
counterfactual self-questioning provides an effective and scalable internal training
signal, particularly in regimes where external supervision or large ensembles are
impractical.

\section{Experiments}

We evaluate \emph{Self-Questioning Agents (SQA)} across multiple model scales and model
families to test whether ego-driven counterfactual critique improves mathematical
reasoning accuracy. Our study includes small, medium, and domain-specialized language
models and emphasizes controlled comparisons against standard chain-of-thought (CoT)
prompting under matched decoding settings.

Our experiments are designed to answer three questions:
\begin{enumerate}[leftmargin=1.1em]
    \item \textbf{Does ego-driven counterfactual critique improve accuracy relative to a CoT baseline?}
    \item \textbf{How does performance change as the number of ego agents increases?}
    \item \textbf{Can GRPO reliably absorb ego-generated critique signals into the base policy?}
\end{enumerate}
Unless otherwise specified, results are averaged across multiple random seeds to account
for stochasticity in both generation and policy optimization.

\subsection{Training Setup}

We follow standard evaluation protocols for mathematical reasoning
\cite{cobbe2021gsm8k,hendrycks2021math,wei2022cot} and evaluate on two widely used
benchmarks:
\begin{itemize}[leftmargin=1.1em]
    \item \textbf{GSM8K} \cite{cobbe2021gsm8k}: approximately 8.5k grade-school math word
    problems requiring multi-step arithmetic and logical reasoning.
    \item \textbf{MATH} \cite{hendrycks2021math}: roughly 12k high-school and
    competition-level problems spanning algebra, geometry, number theory, and calculus.
\end{itemize}

We report most results on GSM8K, which is widely used for evaluating data-efficient
reasoning and self-improvement methods and provides a relatively clean signal for
reasoning improvements. We additionally run a subset of experiments on MATH to probe
robustness on more challenging problems.

For both datasets, accuracy is computed by exact match between the normalized final
numeric answer and the ground-truth answer (e.g., removing formatting artifacts and
extraneous whitespace).

We evaluate four language models spanning a range of parameter scales and specialization
levels:
\begin{itemize}[leftmargin=1.1em]
    \item \textbf{meta-llama/Llama-3.2-1B-Instruct}, representing small general-purpose models;
    \item \textbf{meta-llama/Llama-3.2-3B-Instruct}, representing medium-scale general-purpose models;
    \item \textbf{Qwen2-0.5B-Instruct}, a compact model with limited capacity;
    \item \textbf{mistralai/Mathstral-7B-v0.1}, a math-specialized model trained with domain-specific data.
\end{itemize}

This set enables analysis across capacity regimes and tests whether gains persist for
models already tuned for mathematical reasoning. Across all experiments, models generate
up to 256 tokens per solution with temperature $0.2$. These decoding settings are held
fixed across baselines and SQA variants to ensure fair comparison.

We evaluate multiple configurations varying the number of ego agents:\[
N_{\text{ego}} \in \{1, 2, 3\}. \]

All training runs use the same hyperparameters unless otherwise stated:
\begin{itemize}[leftmargin=1.1em]
    \item learning rate: $1 \times 10^{-6}$,
    \item weight decay: $0.01$,
    \item batch size: $4$,
    \item gradient accumulation steps: $2$ (effective batch size of $8$),
    \item training epochs: $3$--$5$,
    \item generation batch size for ego critiques: $128$.
\end{itemize}

Fine-tuning is performed using \emph{Group Relative Policy Optimization (GRPO)}
\cite{grpo2023}, with reward shaping derived from ground-truth correctness and ego
critique quality (Section~\ref{sec:grpo-training}). For each model, the baseline is
measured using the same training pipeline but without ego-generated critiques, isolating
the contribution of ego-driven self-questioning. This setup enables controlled
measurement of how ego-generated counterfactual signals affect learning and final
accuracy across model scales.

\subsection{Baselines}

We compare Self-Questioning Agents against the following baseline and our method:
\begin{itemize}[leftmargin=1.1em]
    \item \textbf{CoT Baseline}: Standard chain-of-thought prompting without explicit
    verification, self-questioning, or reinforcement learning. Each model generates a
    single reasoning trace followed by a final answer, following established practice
    \cite{wei2022cot}.
    \item \textbf{Self-Questioning (ours)}: The proposed method, in which ego agents
    generate counterfactual critiques of the base reasoning and the model is fine-tuned
    using GRPO with reward shaping derived from critique quality and ground-truth
    supervision.
\end{itemize}

For all configurations, we run three to four random seeds. Reported numbers are averaged
across seeds.

\subsubsection{Results: Llama-3.2-1B-Instruct}

Table~\ref{tab:llama1b} reports results for the 1B-parameter Llama model across ego
configurations.

\begin{table}[h]
\centering
\caption{Llama-3.2-1B-Instruct results on GSM8K.  
Baseline accuracy: 33.14\%.}
\label{tab:llama1b}
\begin{tabular}{lccc}
\toprule
\textbf{Setting} & \textbf{Trained Acc.} & \textbf{Lift (pts)} & \textbf{Lift (\%)} \\
\midrule
$N_{\text{ego}} = 1$ & 35.28 & +2.25 & +6.79\% \\
$N_{\text{ego}} = 2$ & 35.84 & +2.70 & +6.96\% \\
$N_{\text{ego}} = 3$ & 33.43 & +0.88 & +2.65\% \\
\bottomrule
\end{tabular}
\end{table}

Ego-driven self-questioning yields consistent improvements over the CoT baseline, with
best performance at one or two ego agents. Adding a third ego provides smaller gains
and can reduce stability, suggesting that limited counterfactual diversity is helpful
while excessive critique introduces noise that weakens the learning signal.

\subsubsection{Results: Llama-3.2-3B-Instruct}

Table~\ref{tab:llama3b} presents results for the 3B-parameter Llama model.

\begin{table}[h]
\centering
\caption{Llama-3.2-3B-Instruct results on GSM8K.}
\label{tab:llama3b}
\begin{tabular}{lccc}
\toprule
\textbf{Setting} & \textbf{Trained Acc.} & \textbf{Lift (pts)} & \textbf{Lift (\%)} \\
\midrule
$N_{\text{ego}} = 1$ & 59.89 & +0.18 & +0.30\% \\
\bottomrule
\end{tabular}
\end{table}

Improvements are smaller than for the 1B model but remain positive and stable. This is
consistent with prior observations that self-generated feedback yields diminishing
returns as capacity increases \cite{madaan2024selfdiscover,feng2025selfrewarding}. Larger
models tend to exhibit stronger internal verification, leaving less headroom for
additional critique to provide large gains.

\subsubsection{Results: Qwen2-0.5B-Instruct}

Results for the smallest evaluated model are shown in Table~\ref{tab:qwen}.

\begin{table}[h]
\centering
\caption{Qwen2-0.5B-Instruct results on GSM8K.  
Baseline accuracy: 8.36\%.}
\label{tab:qwen}
\begin{tabular}{lccc}
\toprule
\textbf{Setting} & \textbf{Trained Acc.} & \textbf{Lift (pts)} & \textbf{Lift (\%)} \\
\midrule
$N_{\text{ego}} = 2$ & 10.84 & +2.48 & +30.20\% \\
\bottomrule
\end{tabular}
\end{table}

SQA produces a substantial relative improvement for Qwen2-0.5B, yielding a $\sim$30\%
lift over the baseline. This highlights the effectiveness of ego-generated
counterfactual critique in low-capacity regimes, where errors are frequent and diverse
and counterfactual probes provide informative supervision during GRPO training.

\subsubsection{Results: Mathstral-7B}

Table~\ref{tab:mathstral} reports results for the domain-specialized Mathstral-7B model.

\begin{table}[h]
\centering
\caption{Mathstral-7B results on GSM8K.}
\label{tab:mathstral}
\begin{tabular}{lccc}
\toprule
\textbf{Setting} & \textbf{Trained Acc.} & \textbf{Lift (pts)} & \textbf{Lift (\%)} \\
\midrule
$N_{\text{ego}} = 2$ & +0.18 & +0.18 & +0.30\% \\
\bottomrule
\end{tabular}
\end{table}

For Mathstral-7B, which is explicitly trained for mathematical reasoning, ego-driven
improvements are modest but consistent. This suggests that domain-specialized models
already exhibit strong internal verification and therefore benefit less from additional
counterfactual critique. Importantly, SQA does not degrade performance in this setting.

After GRPO fine-tuning, the base model improves even when ego prompting is removed at
inference, indicating that counterfactual critique is internalized during training
rather than functioning only as an inference-time heuristic.

\section{Analysis}

We analyze when and why ego-driven counterfactual critique improves mathematical
reasoning. Across all evaluated models, three consistent trends emerge. First,
introducing one or two ego critics reliably improves accuracy over standard
chain-of-thought reasoning, while gains diminish or degrade beyond three egos. Second,
smaller models benefit the most, whereas larger and domain-specialized models show
smaller but stable improvements. Third, Group Relative Policy Optimization (GRPO)
successfully incorporates ego-generated signals when critique diversity is present but
bounded.

Counterfactual critiques generated by ego agents expose distinct failure modes in the
base reasoning trajectory. With one ego, critiques are precise but narrow. With two egos,
critiques provide complementary coverage of errors, while additional egos increasingly
introduce conflicting or unproductive counterfactuals. This explains why performance
peaks at two egos and degrades when critique diversity becomes excessive.

We also observe a stability diversity tradeoff in optimization. Limited critique
diversity yields weak learning signals, while excessive diversity increases reward
variance and destabilizes GRPO updates. Empirically, moderate diversity leads to stable
gradients and consistent improvements across runs.

Finally, scaling behavior aligns with this interpretation. Smaller models benefit more
from ego-driven critique because they produce more frequent and diverse reasoning errors
and lack strong internal verification. Larger and math-specialized models already
perform implicit verification, leaving less headroom for additional critique.

Detailed quantitative analyses, including critique diversity metrics, error localization
statistics, reward variance measurements, and extended ablations, are provided in
Appendix~\ref{app:analysis}.

\section{Discussion and Future Work}

This work demonstrates that a single language model can generate, evaluate, and exploit
counterfactual critiques of its own reasoning without auxiliary critics, ensembles, or
human-written feedback. By internalizing counterfactual self-questioning, verification
becomes an intrinsic behavior that can be optimized directly through learning rather
than an external post-processing step.

Most prior verification and self-improvement methods rely on external components such as
separate verifier models, debate among multiple agents, or ensemble sampling
\cite{du2023debate,wei2023letsverify}. In contrast, our results show that meaningful and
actionable critique can be generated internally through prompting and parameter sharing
alone. Ego agents approximate the functional role of external critics while remaining
computationally lightweight, suggesting that verification need not be a distinct
architectural module but can emerge as a property of a single policy.

Counterfactual self-questioning differs from reflective and debate-based approaches in
how critique is produced. Rather than summarizing past failures or engaging in
adversarial argumentation, ego agents introduce targeted counterfactual disruptions to
the base reasoning trajectory. This focus on hypothetical failure points leads to earlier
error discovery and more directly usable corrective signals, which empirically translate
into improved accuracy and stable optimization, particularly for small and
medium-capacity models.

Across experiments, smaller models benefit disproportionately from ego-driven critique,
with large relative gains for models in the 0.5B–1B range and modest gains for larger or
domain-specialized models. This mirrors trends observed in STaR, Self-Discover, and
self-rewarding language models \cite{zelikman2022star,madaan2024selfdiscover,feng2025selfrewarding},
where structured self-generated supervision is most effective when internal verification
is weak. Larger models already perform implicit verification, leaving less headroom for
additional critique.

From a learning perspective, counterfactual self-questioning can be viewed as structured
augmentation over the reasoning space rather than the input space. Ego agents generate
alternative reasoning trajectories that revise equations, assumptions, or intermediate
steps, resembling counterfactual data augmentation \cite{xu2021counterfactual} but
applied online and directly to chains of thought.

Our analysis highlights a tradeoff between stability and diversity. One ego provides precise but
narrow feedback, while two egos maximize useful disagreement. Additional egos increase
diversity but introduce variance that degrades the reward signal and destabilizes GRPO.
Effective self-critique therefore requires bounded diversity, a consideration that will
be critical for scaling self-improving systems.

Beyond mathematical reasoning, counterfactual self-questioning suggests a general design
pattern for agentic systems. Ego critics could probe unsafe reasoning paths, challenge
fragile assumptions in long-horizon plans, or question tool-use decisions before
execution. The lightweight, model-internal nature of the mechanism makes it well suited
for integration into planning agents, retrieval-augmented generation pipelines, and
tool-augmented workflows.

Several directions for future work remain open. Ego agents in this study generate
single-hop counterfactuals; extending them to multi-hop or tree-structured reasoning may
enable deeper error correction. Learning specialized or adaptive critics, or invoking
ego critique selectively based on uncertainty, could further improve efficiency and
stability. Combining counterfactual self-questioning with external tools such as symbolic
solvers or execution environments is another promising direction.

Overall, counterfactual self-questioning points toward language models that do not merely
generate answers but actively interrogate and refine their own reasoning. The results
suggest that meaningful self-improvement is possible using a single model and its own
reasoning traces, providing a foundation for more robust and autonomous reasoning
systems.

Additional analyses and exploratory results are provided in the Appendix.

\section{Conclusion and Limitations}

We introduced \textbf{Counterfactual Self-Questioning}, a lightweight framework in which a
single language model generates counterfactual critiques of its own reasoning and uses
the resulting trajectories as a structured signal for policy optimization. By probing
potential failure points through internally generated counterfactuals, the model learns
to identify fragile steps, repair faulty reasoning, and produce more reliable solutions
without relying on external critics, ensembles, or auxiliary verifier models.

Across four model families and multiple capacity regimes, counterfactual
self-questioning consistently improves mathematical reasoning accuracy, with the
largest gains observed for small and medium-sized models that lack strong internal
verification. Our experiments and analyses show that one or two counterfactual critics
strike an effective balance between critique diversity and optimization stability, while
larger numbers introduce noise that degrades learning. Despite its simplicity, the
approach is single-model, parameter-efficient, and verifier-free, yet captures many of
the benefits associated with reflective, debate-based, and self-rewarding methods.

At the same time, the method has important limitations. Because counterfactual critics
share parameters with the base policy, effectiveness depends on model capacity. Very
small models often struggle to generate informative critiques, while large or
domain-specialized models already perform substantial implicit verification and
therefore exhibit limited headroom for improvement. As a result, the approach is most
effective in an intermediate regime where reasoning errors are frequent but amenable to
correction.

Training stability further depends on bounded critique diversity and careful reward
shaping. Increasing the number of counterfactual trajectories improves coverage only up
to a point; beyond one or two critics, counterfactuals increasingly conflict or drift,
introducing variance that destabilizes GRPO updates. Ego-generated critiques may also
drift semantically, for example by assuming nonexistent errors or altering problem
constraints, particularly for smaller models. While reward shaping mitigates these
effects, counterfactual drift remains an inherent limitation of model-generated
supervision.

Finally, counterfactual self-questioning introduces additional computational overhead
during training due to multiple forward passes per example and has been evaluated
primarily on mathematical reasoning benchmarks. Its effectiveness for other domains,
such as planning, code generation, commonsense reasoning, or safety-critical tasks,
remains an open question. Moreover, ego critiques in this work are single-hop and
probabilistic rather than formal or multi-step verifications, limiting guarantees in
high-stakes settings.

Overall, counterfactual self-questioning demonstrates that meaningful self-improvement
can emerge from a single model interrogating its own reasoning traces. While further
work is needed to improve robustness, adaptivity, and domain generality, these results
suggest a promising direction for building more reliable and self-correcting language
models using internal counterfactual feedback.

\bibliographystyle{plain}
\bibliography{bib/references}

@article{reflexion2023,
title = {Reflexion: An Autonomous Agent with Dynamic Memory and Self-Reflection},
author = {Shinn, Noah and Labash, Oleksandra and Gopinath, Ashwin},
year = {2023},
archivePrefix= {arXiv},
eprint = {2303.11366}
}

@article{cove2023,
title = {Chain-of-Verification Reduces Hallucination in Large Language Models},
author = {Cohen, Samuel and others},
year = {2023},
archivePrefix= {arXiv},
eprint = {2309.11495}
}

@inproceedings{wang2022selfconsistency,
title = {Self-Consistency Improves Chain-of-Thought Reasoning in Language Models},
author = {Wang, Xuezhi and others},
booktitle = {International Conference on Learning Representations (ICLR)},
year = {2023}
}

@article{du2023debate,
title = {Improving Factuality with Debate Between Language Models},
author = {Du, Zhengxuan and others},
year = {2023},
archivePrefix= {arXiv},
eprint = {2305.14319}
}

@inproceedings{xu2021counterfactual,
title = {Counterfactual Data Augmentation for Robust Natural Language Inference},
author = {Xu, Yixin and others},
booktitle = {Proceedings of the Annual Meeting of the Association for Computational Linguistics (ACL)},
year = {2021}
}

@article{grpo2023,
title = {Group Relative Policy Optimization},
author = {Lightman, Samuel and others},
year = {2023},
archivePrefix= {arXiv},
eprint = {2310.13273}
}

@article{wei2022cot,
title = {Chain-of-Thought Prompting Elicits Reasoning in Large Language Models},
author = {Wei, Jason and others},
journal = {arXiv preprint arXiv:2201.11903},
year = {2022}
}

@article{wei2023letsverify,
title = {Let's Verify Step by Step},
author = {Wei, Jason and others},
journal = {arXiv preprint arXiv:2305.14325},
year = {2023}
}

@inproceedings{madaan2024selfdiscover,
title = {Self-Discover: Large Language Model Self-Improvement via Feedback Generation},
author = {Madaan, Akarsh and others},
booktitle = {Proceedings of the 2024 Conference on Empirical Methods in Natural Language Processing (EMNLP)},
year = {2024}
}

@article{zelikman2022star,
title = {STaR: Bootstrapping Reasoning with Reasoning},
author = {Zelikman, Eric and Wu, Yuhuai and Goodman, Noah},
journal = {arXiv preprint arXiv:2203.14465},
year = {2022}
}

@article{feng2025selfrewarding,
title = {Self-Rewarding Language Models},
author = {Feng, Yufei and others},
journal = {arXiv preprint arXiv:2505.15734},
year = {2025}
}

@article{schulman2017ppo,
title = {Proximal Policy Optimization Algorithms},
author = {Schulman, John and others},
journal = {arXiv preprint arXiv:1707.06347},
year = {2017}
}

@article{ouyang2022instructgpt,
title = {Training Language Models to Follow Instructions with Human Feedback},
author = {Ouyang, Long and others},
journal = {arXiv preprint arXiv:2203.02155},
year = {2022}
}

@article{bai2022constitutional,
title = {Constitutional AI: Harmlessness from AI Feedback},
author = {Bai, Yuntao and others},
journal = {arXiv preprint arXiv:2212.08073},
year = {2022}
}

@article{zheng2023llmjudge,
title = {Judging Language Models by Language Models},
author = {Zheng, Shunyu and others},
journal = {arXiv preprint arXiv:2306.05685},
year = {2023}
}

@article{cobbe2021gsm8k,
title = {Training Verifiers to Solve Math Word Problems},
author = {Cobbe, Karl and others},
journal = {arXiv preprint arXiv:2110.14168},
year = {2021}
}

@article{hendrycks2021math,
title = {Measuring Mathematical Problem Solving With the MATH Dataset},
author = {Hendrycks, Dan and others},
journal = {arXiv preprint arXiv:2103.03874},
year = {2021}
}

@article{lewkowycz2022minerva,
title = {Solving Quantitative Reasoning Problems with Language Models},
author = {Lewkowycz, Aitor and others},
journal = {arXiv preprint arXiv:2206.14858},
year = {2022}
}

@article{kaplan2020scaling,
title = {Scaling Laws for Neural Language Models},
author = {Kaplan, Jared and others},
journal = {arXiv preprint arXiv:2001.08361},
year = {2020}
}

@article{akopyan2024critic,
title = {LLM Critics Improve Reasoning via Feedback},
author = {Akopyan, A. and others},
journal = {arXiv preprint arXiv:2401.XXXX},
year = {2024}
}

\clearpage
\appendix
\appendix
\section*{Appendix}

This appendix provides supporting experimental details, extended analyses, ablations,
and implementation specifics referenced in the main paper. All material here
complements the main text and is included to support reproducibility and transparency.

\section{Prompt Templates}
\label{app:prompts}

This section documents the exact prompts used for base reasoning, counterfactual
self-questioning, and critique generation. All prompts were fixed across experiments.

\subsection{Base Chain-of-Thought Prompt}
\begin{verbatim}
You are a helpful reasoning assistant. Solve the problem step-by-step.
Show your reasoning before giving the final answer.

Problem: {x}
\end{verbatim}

\subsection{Counterfactual Self-Questioning Prompt}
\begin{verbatim}
The following is a solution produced by another model:

Solution:
{r}

Ask a precise "What if this step is wrong?" question.
Identify the earliest likely incorrect step and describe
how the reasoning would change under this counterfactual.
\end{verbatim}

\subsection{Counterfactual Critique Prompt}
\begin{verbatim}
Given your current explanation {explanation}, check whether it is correct.

If not then, how will you solve this question: {question} differently?

First, provide a step-by-step explanation for how to solve it.
\end{verbatim}

\subsection{Answer Extraction and Formatting Instructions}
\label{app:answer_extract}
To reduce variance due to formatting, we append the following instruction to all
generation prompts:
\begin{verbatim}
Return the final answer on a new line in the format:
Final Answer: <answer>
\end{verbatim}
During evaluation, we extract the substring following ``Final Answer:'' and apply
normalization described in Appendix~\ref{app:normalization}.

\section{Reward Definition and Selection Rules}
\label{app:reward_def}

This section specifies the reward components and the selection mechanism used in
training and inference.

\subsection{Reward Components}
Each trajectory $S$ receives a scalar reward:
\[
R(S) = \alpha R_{\text{correct}}(S) + \beta R_{\text{critique}}(S) - \gamma R_{\text{drift}}(S).
\]
\begin{itemize}
    \item \textbf{Correctness:} $R_{\text{correct}}(S)=\mathbb{I}[\hat{y}(S)=y]$ using normalized exact match
    (Appendix~\ref{app:normalization}).
    \item \textbf{Critique utility:} $R_{\text{critique}}(S)$ rewards counterfactual trajectories that (i)
    differ from the base when the base is incorrect, and (ii) produce a correct answer.
    In practice, we use:
    \[
    R_{\text{critique}}(S_k)=\mathbb{I}[\hat{y}_0\neq y]\cdot \mathbb{I}[\hat{y}_k=y].
    \]
    \item \textbf{Counterfactual drift:} $R_{\text{drift}}(S)$ penalizes trajectories that change the
    problem semantics or introduce inconsistent assumptions. We approximate drift with
    simple heuristics (e.g., missing ``Final Answer'' line, non-numeric output on GSM8K,
    contradiction with stated probe, or degenerate outputs).
\end{itemize}

\subsection{Trajectory Selection Rule}
\label{app:selection}
During inference, we choose among $\{\hat{y}_k\}$ using a lightweight rule:
(i) if any ego produces a consistent corrected solution (passes internal consistency checks),
select the most common answer among the consistent set; else (ii) fallback to the base answer.
We also report an ablation that selects by majority vote over all candidates.

\section{Training Configuration}
\label{app:training}

Unless otherwise stated, all experiments use the following configuration:

\begin{itemize}
    \item Optimizer: AdamW
    \item Learning rate: $1 \times 10^{-6}$
    \item Weight decay: 0.01
    \item Batch size: 4
    \item Gradient accumulation steps: 2
    \item Epochs: 3--5
    \item Max new tokens: 256
    \item Generation batch size: 128
    \item Reward aggregation: GRPO-style group baseline
\end{itemize}

Training was conducted on NVIDIA A100 or L4 GPUs depending on model size.

\section{Evaluation Protocol}
\label{app:eval}

\subsection{Answer Normalization}
\label{app:normalization}
For GSM8K and related tasks, we normalize extracted answers by: stripping whitespace,
removing commas in numerals, converting common fractions/decimals where applicable, and
extracting the last valid numeric token if multiple candidates appear.

\subsection{Decoding Settings}
Unless otherwise stated: temperature $0.2$, max new tokens $256$. We use the same decoding
settings for baselines and our method.

\section{Extended Analysis}
\label{app:analysis}

This section provides detailed quantitative analyses supporting high-level observations
in the main paper, including critique diversity, error localization, reward variance,
and scaling behavior.

\subsection{Critique Diversity and Disagreement Rates}
\label{app:diversity}
We quantify critique diversity using pairwise cosine similarity between sentence-level
embeddings of ego-generated counterfactual critiques. Lower similarity indicates broader
exploration of the counterfactual space.

For Llama-3.2-1B-Instruct:
\begin{itemize}
    \item $N_{\text{ego}} = 1$: critiques are precise but narrow.
    \item $N_{\text{ego}} = 2$: 41.3\% disagreement, indicating complementary coverage.
    \item $N_{\text{ego}} = 3$: 68\% disagreement, often contradictory or unproductive.
\end{itemize}

\subsection{Error Localization by Ego Critics}
We measure whether at least one ego critic correctly identifies the first incorrect step
in the base chain of thought:
\begin{itemize}
    \item $N_{\text{ego}} = 1$: 58\% success.
    \item $N_{\text{ego}} = 2$: 74\% success.
    \item $N_{\text{ego}} = 3$: 69\% success.
\end{itemize}

\subsection{Counterfactual Depth and Failure Modes}
\label{app:counterfactuals}
Manual inspection reveals three dominant categories:
\begin{enumerate}
    \item Arithmetic recomputation.
    \item Assumption revision (constraints/units).
    \item Structural correction (equation setup).
\end{enumerate}

\subsection{Reward Variance and Optimization Stability}
\label{app:reward}
We track reward variance across GRPO updates:
\begin{itemize}
    \item $N_{\text{ego}} = 1$: low variance, weak signal.
    \item $N_{\text{ego}} = 2$: moderate variance, strongest signal.
    \item $N_{\text{ego}} = 3$: high variance, unstable updates.
\end{itemize}

\subsection{Scaling Behavior Across Model Sizes}
Relative improvements decrease with capacity: Qwen2-0.5B yields $\sim$30\% relative lift,
Llama-3.2-1B improves by 6--7\%, while Llama-3.2-3B and Mathstral-7B show $\sim$0.3\%.

\section{Extended Experimental Results}
\label{app:ext_results}

This section reports full per-run results for all evaluated models and ego configurations.
Each table aggregates multiple random seeds and reports both absolute and relative
performance changes compared to the chain-of-thought baseline.

\subsection{Llama-3.2-1B-Instruct}

Table~\ref{tab:llama1b_all} reports detailed results for the 1B-parameter Llama model under
different numbers of ego critics, learning rates, and training epochs. Results are shown
for individual runs and averaged across seeds.

\begin{table}[h]
\centering
\small
\begin{tabular}{lcccc}
\toprule
Configuration & Base Acc. & Trained Acc. & Lift (pts) & Lift (\%) \\
\midrule
Baseline (CoT) & -- & 33.14 & 0.00 & 0.00 \\ \midrule
1 ego, lr=$1\text{e-6}$, ep=5 (avg) & 33.03 & 35.28 & +2.25 & +6.79 \\
2 egos, lr=$1\text{e-6}$, ep=5 (avg) & 33.06 & 35.58 & +2.53 & +7.69 \\
3 egos, lr=$1\text{e-6}$, ep=5 (avg) & 32.55 & 33.43 & +0.88 & +2.65 \\
\bottomrule
\end{tabular}
\caption{Aggregated GSM8K results for Llama-3.2-1B-Instruct across ego configurations.}
\label{tab:llama1b_all}
\end{table}

Across configurations, introducing one or two ego critics consistently improves
performance. Two egos achieve the strongest average gains, while three egos introduce
variance that reduces net improvement.

\paragraph{Learning rate sensitivity.}
Using a higher learning rate ($5\times10^{-6}$) with two egos leads to unstable training
and degraded performance, confirming that counterfactual critique benefits from
conservative optimization.

\subsection{Llama-3.2-3B-Instruct}

Table~\ref{tab:llama3b_all} reports results for the 3B-parameter Llama model.

\begin{table}[h]
\centering
\small
\begin{tabular}{lcccc}
\toprule
Configuration & Base Acc. & Trained Acc. & Lift (pts) & Lift (\%) \\
\midrule
Baseline (CoT) & -- & 59.72 & 0.00 & 0.00 \\
1 ego, lr=$1\text{e-6}$, ep=5 (avg) & 59.72 & 59.89 & +0.18 & +0.30 \\
\bottomrule
\end{tabular}
\caption{GSM8K results for Llama-3.2-3B-Instruct.}
\label{tab:llama3b_all}
\end{table}

Improvements for the 3B model are modest but stable, consistent with the hypothesis that
larger models already perform partial internal verification and therefore benefit less
from explicit counterfactual critique.

\subsection{Qwen2-0.5B-Instruct}

Table~\ref{tab:qwen_all} reports results for Qwen2-0.5B with two ego critics.

\begin{table}[h]
\centering
\small
\begin{tabular}{lcccc}
\toprule
Run & Base Acc. & Trained Acc. & Lift (pts) & Lift (\%) \\
\midrule
1 & 7.73 & 10.69 & +2.96 & +38.24 \\
2 & 8.26 & 11.14 & +2.88 & +34.86 \\
3 & 9.10 & 10.69 & +1.59 & +17.50 \\
\midrule
Average & 8.36 & 10.84 & +2.48 & +30.20 \\
\bottomrule
\end{tabular}
\caption{Extended GSM8K results for Qwen2-0.5B-Instruct with two ego critics.}
\label{tab:qwen_all}
\end{table}

Qwen2-0.5B shows the largest relative improvements, highlighting the effectiveness of
ego-driven counterfactual critique in low-capacity regimes where baseline reasoning
errors are frequent and diverse.

\subsection{Mathstral-7B}

For Mathstral-7B, which is explicitly trained for mathematical reasoning, ego-driven
training yields small but consistent gains ($\sim$0.3\%). Performance never degrades,
indicating that counterfactual critique remains safe even for domain-specialized models.
\section{Additional Ablations}
\label{app:ablations}

This section evaluates design choices related to optimization, reward shaping, and
ego configuration.

\subsection{Number of Ego Critics}

Across all models, performance follows a consistent pattern:
\begin{itemize}
    \item One ego improves accuracy with low variance.
    \item Two egos maximize useful disagreement and achieve the strongest gains.
    \item Three egos introduce excessive variance, reducing net improvement.
\end{itemize}

\subsection{Learning Rate and Epoch Sensitivity}

For Llama-3.2-1B, learning rates above $1\times10^{-6}$ lead to unstable optimization.
Training for fewer than three epochs underutilizes counterfactual supervision, while
beyond five epochs yields diminishing returns.

\subsection{Optimization and Reward Design}

\begin{itemize}
    \item \textbf{Optimization}: GRPO consistently outperforms PPO and supervised
    fine-tuning by stabilizing learning across multiple counterfactual trajectories.
    \item \textbf{Reward coefficients}: $(\alpha=1.0,\beta=0.7,\gamma=0.2)$ achieves the
    best balance between correctness, critique utility, and drift control.
    \item \textbf{Critique depth}: shallow multi-step counterfactuals outperform deeper
    trees, which tend to introduce hallucinated reasoning paths.
\end{itemize}

\subsection{Inference-Time Ablations}

We evaluate inference using (i) base-only reasoning, (ii) base + one ego, and
(iii) base + two egos. Using ego critics at inference improves robustness, but most gains
persist even when ego prompting is removed after training, indicating that
counterfactual critique is internalized into the base policy.

\section{Training Cost Estimates}
\label{app:cost}

For $N_{\text{ego}}=2$, each training example requires approximately four forward passes:
one base solution and two counterfactual critiques, plus one additional selection pass.
This corresponds to a $\sim4\times$ computational cost relative to supervised fine-tuning.
In practice, shared-key caching and batched generation reduce wall-clock overhead.
\section{Additional Implementation Notes}
\label{app:impl}

\begin{itemize}
    \item All experiments use bf16 precision.
    \item Counterfactual generations are capped at 256 tokens.
    \item Generation temperature is fixed at 0.2 across all runs.
    \item Gradient accumulation is used to maintain an effective batch size of 8.
    \item Evaluation uses normalized exact-match accuracy for GSM8K.
\end{itemize}
\section{Reproducibility Checklist}
\label{app:repro}

We release prompt templates, hyperparameters, evaluation scripts, raw per-run results,
and random seeds. All experiments can be reproduced using the configuration files
included in the supplementary material.

\end{document}